# Locally Grouped and Scale-Guided Attention for Dense Pest Counting

Chang-Hwan Son

Department of Software Science & Engineering, Kunsan National University

558 Daehak-ro, Gunsan-si 54150, Republic of Korea

*Corresponding Author

Phone Number: 82-63-469-8915; Fax Number: 82-63-469-7432

E-MAIL: changhwan76.son@gmail.com; cson@kunsan.ac.kr

**Abstract**

This study introduces a *new dense pest counting problem* to predict densely distributed pests captured by digital traps. Unlike traditional detection-based counting models for sparsely distributed objects, trap-based pest counting must deal with dense pest distributions that pose challenges such as severe occlusion, wide pose variation, and similar appearances in colors and textures. To address these problems, it is essential to incorporate the local attention mechanism, which identifies locally important and unimportant areas to learn locally grouped features, thereby enhancing discriminative performance. Accordingly, this study presents a novel design that integrates locally grouped and scale-guided attention into a multiscale CenterNet framework. To group local features with similar attributes, a straightforward method is introduced using the heatmap predicted by the first hourglass containing pest centroid information, which eliminates the need for complex clustering models. To enhance attentiveness, the pixel attention module transforms the heatmap into a learnable map. Subsequently, scale-guided attention is deployed to make the object and background features more discriminative, achieving multiscale feature fusion. Through experiments, the proposed model is verified to enhance object features based on local grouping and discriminative feature attention learning. Additionally, the proposed model is highly effective in overcoming occlusion and pose variation problems, making it more suitable for dense pest counting. *In particular, the proposed model outperforms state-of-the-art models by a large margin, with a remarkable contribution to dense pest counting. To be specific, the proposed model can improve accuracy by approximately 1058.2%, 90.8%, and 31.3% in terms of mean absolute error (MAE) compared with Faster RCNN, YOLOv7, and multiscale CenterNet, respectively, for a uniform dataset.*

## 1. Introduction

Pest outbreaks severely damage crop leaves and fruits, reducing product quality and causing economic losses. According to the Food and Agriculture Organization, pests incur an estimated 40% annual loss in global crop production [1], necessitating timely pest control measures. One such measure involves digital traps installed in an open field, with light sources and pheromones used to attract and trap pests, depending on the pest type. The captured pests are photographed using a built-in camera, and machine learning algorithms are deployed to predict the number of pests. Early attempts to install such traps employed handcrafted feature extraction and shallow classifiers, which resulted in low pest counting accuracy. However, there has been a recent shift toward applying deep learning to digital trapping.

### 1.1. Issues in trap-based pest counting

Popular detection-based counting models, including Faster R-CNN [2] and YOLO [3], predict bounding boxes and sum them to calculate the number of objects in an image. These object counting models primarily target sparsely distributed objects such as cars and pedestrians. That is, the objects are separated from each other, and the number of objects is typically less than a few dozen. Additionally, these objects tend to be sufficiently large to be detected by a bounding box. In contrast, trap-based pest counting has major limitations, as shown in

Fig. 1. First, the number of pests may vary significantly depending on the season, with up to 200 pests accommodated by digital traps. Second, occlusion becomes a serious problem when many pests are caught in digital traps, as pests often adhere together, becoming obscured by other pests. Third, the targets are relatively small and similar in appearance, making them difficult to distinguish. Through experiments, we demonstrated that conventional detection-based counting models perform significantly worse than expected when handling densely distributed pests, indicating that these models are not suitable for trap-based pest counting. This necessitates the development of a new counting model optimized for *densely distributed and clustered pests*. **Throughout this study, we refer to this new counting problem as *dense pest counting*.**

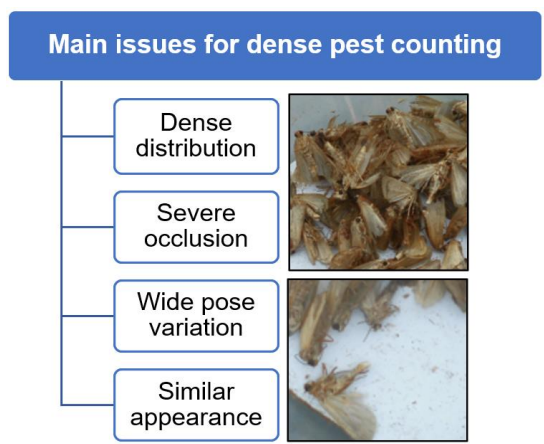

Fig. 1. Dense pest distributions pose challenges such as severe occlusion, shape distortion, pose variability, and similar appearance in colors and textures, which hinder traditional detection-based counting models from distinguishing the pests.

**1.2. Proposed approach**

In this study, we present a new counting model that targets densely distributed pests captured using digital traps. To address challenges such as occlusion and similar appearances, it is crucial to implement local attention mechanisms that determine locally important and unimportant areas to learn discriminative features. However, this objective encompasses complex tasks such as clustering and foreground separation, which may be difficult or even infeasible in the context of dense pest distributions. To simplify these tasks, we introduce a simple yet effective approach to leverage the heatmap information predicted by the first hourglass, which is a stacked backbone in a multiscale CenterNet framework [4]. The heatmap appears as a black-and-white map, where white colors indicate the centroids of pests (also called *keypoints*) and black colors represent background areas. Therefore, this heatmap serves as a clustering map, serving as a strong foundation for generating a learnable attention map.

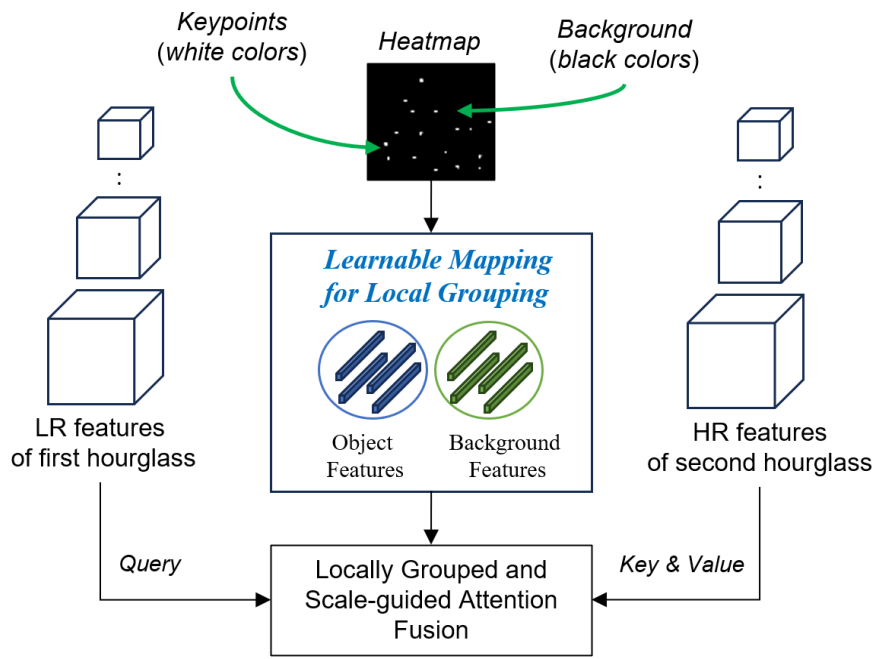

Fig. 2. Proposed approach based on locally grouped and scale-guided attention for dense pest counting.

A framework of the proposed method is presented in Fig. 2. The underlying idea is to separate keypoints (i.e., the centroids of pests) from the background and boost object features while suppressing the background features. First, to group local features by learnable mapping, the heatmap serves as a starting point to indicate which features are objects or backgrounds and is then transformed into a learnable attention map. Next, locally grouped and scale-guided attention are applied to multiscale low-resolution (LR) and high-resolution (HR) features extracted from stacked backbones. The LR object features filtered by the learnable attention map can be used to determine which HR features are emphasized. That is, the LR and HR object features are used as queries (Q) and keys (K), respectively, for HR object feature filtering and fusion. This novel form of attention, henceforth referred to as locally grouped and scale-guided attention (LGSA), can enhance the discrimination between the object and background features owing to local grouping and attention learning, eventually yielding a significant improvement in dense pest counting.

In this study, the multiscale CenterNet framework [4] was selected as the baseline for two key reasons. First, conventional detection-based counting models produce small tensor outputs for bounding box prediction, which may be undesirable for dense pests because object location information is unavoidably lost. In contrast, multiscale CenterNet, which is an anchor-free detector, produces a relatively high-resolution tensor output, which is beneficial for detecting densely distributed pests. Furthermore, CenterNet includes stacked backbones. Therefore, the heat map information predicted by the first hourglass can serve as a good initial point for local feature grouping, making it easier to realize the proposed LGSA method.

The main contribution of this study is a novel pest-counting model optimized for densely distributed pests. In particular, we introduce a simple yet effective approach for enabling local feature grouping within the framework of a multiscale CenterNet architecture. We also describe the design of locally grouped and scale-guided attention to enhance object features and suppress background features in detail. Our experimental results support the assumption that the proposed LGSA model is better suited to handling densely distributed pests than conventional detection models. ***Surprisingly, the proposed model outperformed state-of-the-art models by a large margin, which is a remarkable contribution to dense pest counting. Furthermore, owing to local attention learning, the proposed LGSA model was verified to be highly effective in overcoming challenges such as occlusion and pose variation.***

## 2. Related Works

Pest detection methods can be categorized as trap-based [4,1] and non-trap-based [5-8]. Trap-based methods localize pests and identify the pest type from input images captured by a digital trap. Because it is impossible to count all pests in a field, digital traps are being adopted to statistically determine the number of pests present in the field. Digital traps can lure pests with light sources or pheromones and photograph them using built-in cameras. Because each trap can only be used to attract one species, trap-based pest detection focuses solely on predicting the number of pests. In contrast, non-trap-based pest detection methods [5-8] typically employ refined images collected from the Internet, each of which represents one or two pests. Therefore, these methods are more concerned with identifying rather than counting pests.

**2.1 Non-trap-based pest detection and recognition**

General object detectors, such as those in the R-CNN [2] and YOLO [3] families, can be directly used for pest detection. A simple approach is to apply fine-tuning to fit model parameters to a new pest dataset. Variants of these methods have been developed to overcome challenges such as extremely small object sizes [9] and long-tailed distributions [10]. For instance, pests such as rice planthoppers and wheat aphids are too small to be detected by general object detectors. To address this problem, ClusRPN [9] was proposed to output the location candidates of cluster regions and apply local detectors separately. Although this model is effective for extremely small-scale wheels, its implementation appears complex and its performance is sensitive to the object density level. As a handheld pest-monitoring system [10], the rice planthopper search network (RPSN) was proposed to extract multiple high-quality proposal regions from large-scale pest images with tiny objects. Additionally, a sensitive score matrix was employed to further enhance performance in terms of class prediction and bounding box regression. However, the RPSN architecture was adopted from that of Faster R-CNN [2], with the two models exhibiting only slight differences in performance. Additional multiscale feature extraction methods based on dilated convolution [12] and multiple branches [13] with different kernel sizes have been introduced to generate multiple scales and adjust receptive fields for pest recognition. These are simple yet effective models that require only convolutional filters. Whereas all aforementioned models follow the convolutional neural network (CNN) framework, vision transformers (ViTs) have recently been introduced as an alternative architecture. In most cases, conventional ViTs [14,15] are used directly. Variants of ViTs have also been developed for multilabel classification [16], multiscale patch fusion [17], and region-of-interest (ROI) attention [18].

Furthermore, pest image datasets may exhibit long-tailed distributions [10] that vary with respect to period and region. Such a distribution poses a challenge in avoiding biases toward the head classes and determining an optimal boundary decision for the tail classes. A simple strategy involves generating synthetic images using data augmentation techniques that can be implemented using traditional image enhancement and deep generative models. Image enhancement methods adjust the image appearance – such as contrast, edges, and colors – whereas deep generative models predict real data distributions to generate high-quality samples. Generative adversarial networks (GANs) are popular tools for learning the distributions of datasets through adversarial training. Although there are more sophisticated approaches such as confidence loss design[19], meta-learning[20], and decoupled learning[21], data augmentation techniques remain mainstream owing to their simplicity and efficiency.

**2.2 Trap-based pest detection and counting**

Unlike non-trap-based pest detection methods, trap-based pest detection and counting remains a scarcely researched topic. One early trap-based model [22] moves a sliding window to the input image and trains a classifier for moth detection. Although this approach is the first attempt to apply a deep CNN for this task, it dualizes feature extraction and classifier learning; in other words, it does not achieve end-to-end learning. More recently, PestNet [23] has been proposed for large-scale multiclass pest detection. Within the PestNet architecture, spatial and channel attention are fused into a ResNet backbone to enhance feature discrimination, and the RPN is slightly upgraded with a position-sensitive score map to encode positional information. However, because the comparative models were restricted to SDD and Faster RCNN, PestNet can be expected to underperform compared to the latest object detection models. Pest-YOLO [19] and AgriPest-YOLO [1] are two recent trap-based pest detection networks. Pest-YOLO [19] improves upon YOLOv4 by modeling a new confidence loss, which is a variant of focal loss, to focus on more hard samples while reducing the weights of easily classified samples. Furthermore, a confluence strategy was introduced to optimize the selection of candidate bounding boxes. Consequently, Pest-YOLO slightly outperformed the standard YOLOv4 model. AgriPest-YOLO[1] is based on YOLOv5s and includes a new coordinate attention module that contains both horizontal and vertical directional awareness information, with grouping spatial pyramid pooling fast (GSPPF) modules added to intensify multiscale pest feature extraction. PestNet, Pest-YOLO, and AgriPest-YOLO all employ the RCNN and YOLO families as baselines, representing anchor-based object detectors. However, these detectors might not be optimal for dense objects, as images of such objects tend to exhibit large overlaps and occlusions. Specifically, anchor-based detectors generally output small tensors for bounding box prediction, which are insufficient to represent the locations of densely distributed objects. Therefore, a more advanced model optimized for dense pests must be developed.

Unlike natural images that include pedestrians and cars, images of pests contain tens to hundreds of objects of interest. When the objective is strictly pest-counting, the application of crowd-counting techniques can offer viable solutions. Crowd counting can be used to predict the density map, which is then summed to obtain the pest count. Starting with CrowdNet [24], which was the first attempt in the field to apply a deep CNN, multiscale architectures such as MCNN[25], SANet[26], and ICCNet[27] – which utilize different kernels and multiple branches – were introduced to overcome the scale problem. Another line of research involves learning optimal kernels [28] and modifying the loss function based on optimal transport [29] to generate ground-truth density maps. We conducted experiments to determine whether these crowd-counting approaches can work well for dense pest counting and whether their performance is comparable to that of object detection models. The related details are discussed in the Experimental Results section.

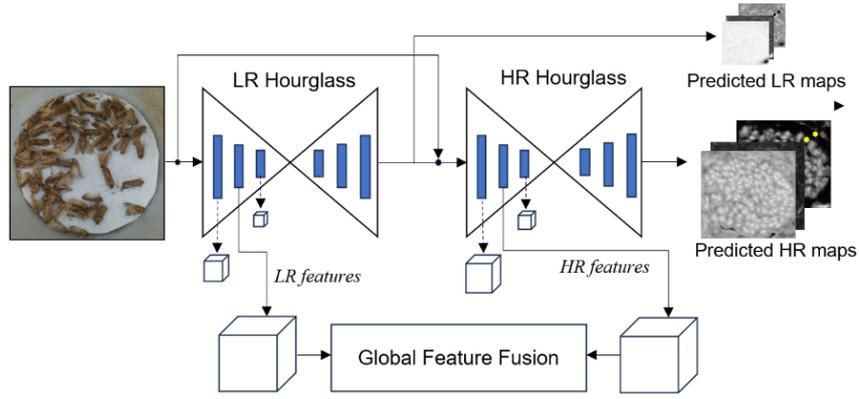

Fig. 3. Conventional multiscale CenterNet for dense pest counting.

## 3. Overview of multiscale CenterNet

Recently, the multiscale CenterNet [4] was developed for dense pest counting, demonstrating powerful performance owing to multiscale LR and HR fusion based on global attention. Fig. 3 illustrates the core modules of multiscale CenterNet, which contain two hourglasses with different scales, referred to as LR and HR hourglasses. This multiscale design can generate high-resolution prediction maps – including heat, offset, and bounding box maps – to be used for the final pest detection. These prediction maps can reduce sampling errors and make pest detectors more robust to occlusion problems. Moreover, the LR feature information is leveraged to effectively learn the HR features. That is, the LR features flow into the HR hourglass via a between-hourglass skip connection, and multiscale fusion is performed based on scale-dot product attention.

$$\boldsymbol{Q} = \boldsymbol{V} = \mathcal{M}\left(\mathcal{H}_{HR}^{\ell}\right) \qquad (1)$$

$$\boldsymbol{K} = \mathcal{M}\left(\mathcal{H}_{LR}^{\ell}\right) \qquad (2)$$

$$\text{Atten}(\boldsymbol{Q}, \boldsymbol{K}, \boldsymbol{V}) = softmax\left(\frac{\boldsymbol{Q}\boldsymbol{K}^T}{\sqrt{d_k}}\right)\boldsymbol{V} \qquad (3)$$

Here, $\mathcal{H}_{HR}^{\ell}$ and $\mathcal{H}_{LR}^{\ell}$ indicate output feature maps at the $\ell$-th residual block in the HR and LR hourglasses, respectively, and $\mathcal{M}$ is a linear transformation for feature embedding. In Eqs. (1) and (2), the HR features are used as queries ($Q$) and values ($V$), and the LR features are used as queries ($K$). Eq. (3) computes the scale-dot product attention, where $d_k$ denotes the dimension of $K$ and $softmax$ represents the softmax function used to squish the input values into [0-1]. Here, $QK^T$ is a similarity matrix containing the weighted values required for the linear combination of $V$. This equation indicates that the LR features guide HR features by enhancing their focus and precision. Overall, the multiscale attention fusion boosts HR features, thereby improving pest counting accuracy. For reference, $\mathcal{H}_{HR}^{\ell}$ and $\mathcal{H}_{LR}^{\ell}$ have different image scales. In multiscale fusion, upsampling is required to match these scales, or different residual blocks in the HR and LR hourglasses can be set to the same scale.

## 4. Proposed locally grouped and scale-guided attention for dense pest counting

Although the multiscale CenterNet represents a new model optimized for dense pest counting, there is still considerable room for improvement in terms of accuracy. First, the model adopts a global attention approach without local grouping, which is insufficient for distinguishing locally important areas from the background in the *dense feature space.* Therefore, this approach has limitations in localizing densely distributed pests. For dense pest counting, the foreground must be separated from the background, and the keypoints must be distinguishable. Dense pest detection can be further improved by incorporating local grouping and guided attention to learn where to pay attention to the feature maps. To this end, locally grouped and scale-guided attention have been proposed. Furthermore, a pest image dataset with a uniform distribution of pests must be used for evaluation, whereas the multiscale CenterNet has only been tested on a nonuniform pest dataset. Specifically, images with pest counts between 1 and 35 accounted for 80% of the dataset in question, whereas images with pest counts above 100 were rare. Consequently, the pest-counting model was biased toward peak distribution, making its predictive capabilities unclear for an arbitrary number of pests. To verify the generalized counting capability, a uniform pest dataset was tested. To our knowledge, no local attention model has fully addressed challenges like severe occlusion, wide pose variation, and similar appearances in color and texture.

In addition, in-depth analyses and evaluation comparisons of both uniform and non-uniform datasets remain scarce. Motivated by this, our proposed approach was formulated to leverage heatmap information and incorporate a novel architectural design that integrates LGSA into the multiscale CenterNet.

Fig. 4 illustrates the architecture of the proposed LGSA for dense pest counting. The objective of the LGSA is to transform HR features into a more effective form for discriminative feature learning. To achieve this, we propose a simple yet effective approach that leverages the LR heatmap provided by the first hourglass. This eliminates the need for complex tasks such as feature clustering and foreground separation for local grouping, as the LR heatmap contains keypoints (i.e., the centroids of pests). More precisely, in addition to the LR heatmap, the LR bounding box and offset maps are also encompassed by LGSA. As illustrated at the top of Fig. 4, the predicted LR heatmap is close to a grayscale image, although its ground truth is depicted in black and white. Furthermore, LGSA requires LR features extracted from the LR hourglass and prediction maps (i.e., heatmap, bounding box map, and offset map) as side information. The LR heatmap contains information on local grouping, and the LR features represent a good initial solution for pest detection. Thus, this information can provide useful tips on designing the LGSA and controlling the input HR features. In other words, the local grouping information is distilled through the proposed LGSA to make the object HR features more discernible while suppressing the background HR features. Although the proposed LGSA encompasses three maps, we explain the local grouping in terms of the heatmap to provide the readers with an intuitive interpretation of the LGSA. The corresponding details are described below.

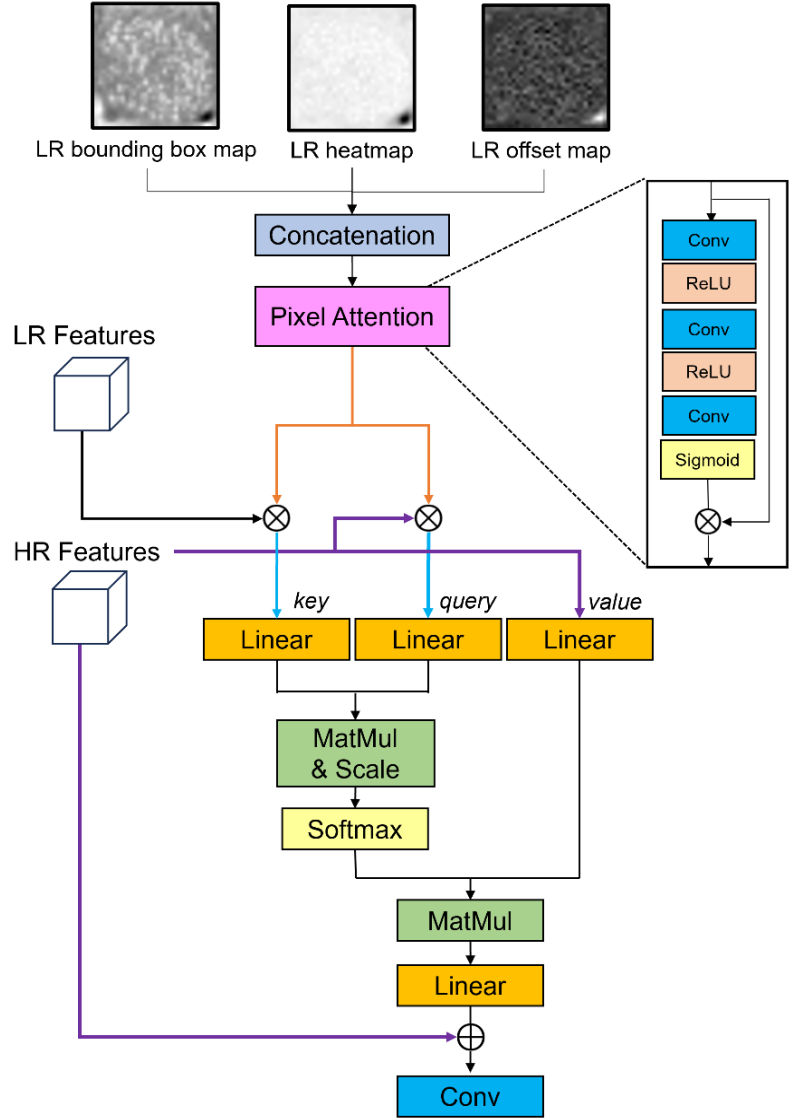

Fig. 4. Proposed locally grouped and scale-guided attention for dense pest counting.

### 4.1. Heatmap design

The LR prediction maps shown at the top of Fig. 4 serve as a good initial point to determine which features belong to the object and background regions. In particular, the heat map contains keypoints (i.e., the centroids of pests) that can be used to straightforwardly categorize features as objects and background features for local grouping. We explain the LGSA in terms of a heatmap to provide readers with an intuitive understanding of the concept of local grouping. For heatmap generation, an annotated map is constructed by marking the centroids of pests in white and all other areas in black. The bounding box labeling task is a prerequisite for calculating the centroids of pests.

$$\mathbf{D} \in [0,1]^{(W/R) \times (H/R) \times C} \quad (4)$$

Here, $W$ and $H$ represent the width and height of an input pest image, respectively, $R$ denotes the stride used to determine the resolution of the heatmap, and $C$ represents the number of object classes. Thus, $\mathbf{D}$ represents the annotated dot map in the form of a sparse binary mask, where each pest is marked with a white dot on its centroid. Subsequently, Gaussian filtering is performed as follows:

$$\mathbf{Y} = \mathbf{D} \odot \mathbf{G} \quad (5)$$

Here, $\mathbf{G}$ is the Gaussian filter and $\odot$ denotes the convolution operator. Because $\mathbf{D}$ is a binary mask, Eq. (2) is equivalent to placing a Gaussian filter on each dot annotation. In the final heat map, the keypoints remain white. In other words, the centroids remain unchanged after Gaussian filtering, which facilitates the detection of maximal keypoints. Given a heatmap $\mathbf{Y}$, the bounding box and offset maps are generated based on the bounding box labeling results. The bounding box map contains the width and height of the ground-truth bounding box at the keypoint, and the offset map contains offsets for 2D pixel coordinates caused by downsampling at the ratio of $R$, used in Eq. (4).

**4.2. LR heatmap learning and prediction**

The three types of prediction maps – heatmap, bounding box map, and offset map – are output by the LR hourglass in the multiscale CenterNet. To train the LR hourglass, the total loss is defined as follows:

$$\mathcal{L}_T^{LR} = \mathcal{L}_h^{LR} + \lambda_b \mathcal{L}_b^{LR} + \lambda_o \mathcal{L}_o^{LR} \quad (6)$$

$$\mathcal{L}_h^{LR} = -\frac{1}{N}\Sigma_k^N \begin{cases} \left(1-\widehat{\mathbf{Y}}^{LR}(k)\right)^\alpha log\left(\widehat{\mathbf{Y}}^{LR}(k)\right) \; if \; \mathbf{Y}^{LR}(k)=1 \\ \left(1-\widehat{\mathbf{Y}}^{LR}(k)\right)^\beta \left(\widehat{\mathbf{Y}}^{LR}(k)\right)^\alpha log\left(1-\widehat{\mathbf{Y}}^{LR}(k)\right) \; otherwise \end{cases} \quad (7)$$

$$\mathcal{L}_b^{LR} = \frac{1}{N}\Sigma_{k=1}^N |\widehat{\mathbf{S}}_k^{LR} - \mathbf{S}_k^{LR}| \quad (8)$$

$$\mathcal{L}_o^{LR} = \frac{1}{N}\Sigma_k^N \left|\widehat{\mathbf{O}}_{\tilde{k}}^{LR} - \left(\frac{k}{R} - \tilde{k}\right)\right|, \; where \; \tilde{k} = \left[\frac{k}{R}\right]. \quad (9)$$

Here, $\mathcal{L}_h^{LR}$, $\mathcal{L}_b^{LR}$, and $\mathcal{L}_o^{LR}$ represent prediction errors for the LR ground-truth heatmap, LR bounding box map, and LR offset map, respectively. First, the LR heatmap loss is modeled based on focal loss to solve the class imbalance problem. This is expressed by Eq. (7), where $\widehat{\mathbf{Y}}^{LR}$ denotes the predicted LR heatmap, the prediction $\mathbf{Y}^{LR}(k)=1$ corresponds to the $k$-th keypoint in the ground-truth LR heatmap, and $N$ denotes the total number of keypoints. Second, in Eq. (8), $\mathbf{S}_k^{LR}$ contains the width and height of the LR bounding box at the $k$-th keypoint, and $\widehat{\mathbf{S}}_k^{LR}$ represents the predicted bounding box map of the same size as $\widehat{\mathbf{Y}}^{LR}$, but with two channels. Thus, $\mathcal{L}_b^{LR}$ is the mean of the absolute differences between the predicted and ground-truth bounding box sizes. Third, the LR offset loss $\mathcal{L}_o^{LR}$ reflects the discretization errors caused by downsampling at a ratio of $R$. In Eq. (9), the parentheses imply rounding off to obtain an integer pixel location, and $\widehat{\mathbf{O}}_{\tilde{k}}^{LR}$ denotes the offset map that has the same size as $\mathbf{S}_k^{LR}$ and contains offsets for the 2D pixel coordinates. In Eq. (6), $\lambda_b$ and $\lambda_o$ denote weights that are set to 1 and 0.1, respectively, whereas in Eq. (7), $\alpha$ and $\beta$ are set to 2 and 4, respectively. Gradient-based optimizers [30] are used to train the total loss.

### 4.3. Pixel attention for learnable heatmap transformation

The LR heatmap is a local grouping result that can separate the object and background features. However, because its distribution of object centroids is sparse, it does not represent an optimal spatial attention map. A more effective approach entails transforming the heat map into a learnable map. For this task, we considered the pixel attention module (PAM) illustrated on the upper-right of Fig. 4, consisting of three types of layers:

convolution (**Conv**), rectified linear unit (**RELU**), and **Sigmoid** for learnable heatmap transformation.

$$\boldsymbol{F} = Concat(\hat{\boldsymbol{Y}}^{LR}, \hat{\boldsymbol{S}}^{LR}, \hat{\boldsymbol{O}}^{LR}) \tag{10}$$

$$\boldsymbol{W} = f^s \circ f^c \circ f^a \circ f^c \circ f^a \circ f^c(\boldsymbol{F}) \tag{11}$$

$$\boldsymbol{M} = \boldsymbol{F} \otimes \boldsymbol{W} \tag{12}$$

Here, $Concat$ is the concatenation layer and $f^s$, $f^c$, and $f^a$ denote the Sigmoid, Conv, and ReLU layers, respectively. The symbols $\circ$ and $\otimes$ represent the composite function and elementwise multiplication operations, respectively. Eq. (10) indicates that the input $\boldsymbol{F}$ of the PAM consists of three prediction maps stacked along the channel axis using a concatenation layer. In Eq. (11), a series of composite functions eventually produces a two-dimensional weighting map $\boldsymbol{W}$ to control which features are emphasized and shrunk. In Eq. (11), the final layer is the sigmoid layer, which squishes values within the range of 0 and 1. The final output of the PAM is the element-wise multiplication of the input $\boldsymbol{F}$ and weighting map $\boldsymbol{W}$, as shown in Eq. (12). This equation indicates that the heatmap, which is the starting point for local feature grouping, can vary spatially according to the learned weighting map $\boldsymbol{W}$. The proposed learnable heatmap transformation enhances the flexibility and accuracy of local feature grouping.

**4.4. Locally grouped and scale-guided attention**

For dense pest counting, it is crucial to separate object features from background features in a dense feature space. Attention models [31] are optimal tools for accomplishing this goal owing to their effectiveness in extracting semantic information and reducing the influence of backgrounds. Early attention models, such as SENet [32] and CBAM [33], focused on designing spatial and channel attention maps within CNN frameworks. More recently, the ViT [14,15] has replaced the CNN framework in various vision tasks as a convolution-free model that relies only on a self-attention mechanism to draw global dependencies. In particular, multihead attention (MHA), a core component of the ViT, can model long-range dependencies between input tokens, exhibiting great flexibility. Accordingly, MHA [31] and its variants [15] have attracted considerable attention in the computer vision field.

In the multiscale CenterNet [4], MHA is used based on a scale-dot product for multiscale LR and HR feature fusion. However, the global attention model has limitations in localizing densely distributed pests. In other words, it may be difficult to distinguish and localize pests in densely distributed areas. To address this challenge, we developed a novel local attention system called LGSA that optimizes dense pest counting. LGSA can be regarded as a modified version of the MHA that can model local feature grouping and boost object features while suppressing background features. As depicted in Fig. 4, LGSA was designed based on the learnable heatmap $M$, which is itself a local grouping result. Thus, it is possible to determine the features of objects and backgrounds based on a learnable heatmap. To complete the LGSA, we adopted a filtering concept represented by a multiplication operation in the frequency domain. Specifically, the learnable heatmap is multiplied by the input LR and HR features to filter out background features.

$$\boldsymbol{Q_F} = \mathcal{H}_{HR}^{\ell} \otimes f^{d(\ell)}(\boldsymbol{M}), \qquad \boldsymbol{K_F} = \mathcal{H}_{LR}^{\ell} \otimes f^{d(\ell)}(\boldsymbol{M}), \qquad \boldsymbol{V} = \mathcal{H}_{HR}^{\ell} \qquad (13)$$

Here, $\mathcal{H}_{HR}^{\ell}$ and $\mathcal{H}_{LR}^{\ell}$ represent the HR and LR features, respectively, extracted from corresponding hourglasses in the $\ell$th residual block. For brevity, the feature embedding $\mathcal{M}$ is omitted in Eq. (13). $f^{d(\ell)}$ denotes the downsampling layer, where the sampling rate depends on the $\ell$-th block. For this stage, we adopted bilinear interpolation. Given the HR and LR features, it is difficult to determine the object features and make them more discriminative; however, this can be facilitated by a learnable heatmap. In Eq. (13), the learnable heatmap $M$ is downsampled and then element-wise multiplied with the HR and LR features to extract the object features, acting as an image filter. In addition, unlike the original MHA [31], the proposed LGSA has different forms for the query, key, and value for scale-guided attention, as shown in Eq. (13). Specifically, the filtered results of the HR and LR features are used as $Q_F$ and $K_F$, whereas the HR features were used as $V$. The subscript $F$ indicates filtering.

$$SIM = Q_F K_F^T \qquad (14)$$

$$LGSA(Q_F, K_F, V) = softmax\left(\frac{SIM}{\sqrt{d_k}}\right) V \qquad (15)$$

Here, $SIM$ is the similarity matrix used to measure the extent to which filtered HR features are similar to those of filtered LR images. Each row of the similarity matrix stores weighting values to apply a linear combination of $V$. The weighted values are normalized using the softmax function. Here, the filtered multiscale features are likely object features, as they belong to one of two local groups. For ease of understanding, we will now consider one of the two groups as an object group. **Because the similarity matrix is the inner product result between object features at different scales, locally grouped attention can be achieved. Fig. 5 shows that the similarity matrix can distinguish between object and background features based on the filtering operation expressed in Eq. (13), which is different from the standard MHA [31]. Moreover, the LR object features filtered by the learnable heatmap can teach the model which HR features are more emphasized, thereby achieving scale-guided attention. In this respect, the proposed attention can be considered simultaneously locally grouped and scale-guided.**

$$\begin{bmatrix} O\ O\ BB\ ,..,\ O \\ SIM \end{bmatrix} \quad \begin{array}{l} O : \text{Object} \\ B : \text{Background} \end{array}$$

Fig. 5. Similarity matrix for local grouping and attention.

The final HR features are updated via skip connection:

$$\mathcal{H}_{HR}^{\ell+1} = f^c\left(\mathcal{H}_{HR}^{\ell} + LGSA(\boldsymbol{Q}_F, \boldsymbol{K}_F, \boldsymbol{V})\right) \tag{16}$$

The updated HR features pass through the convolution layer and are fed into the next layer of the HR hourglass. This process is repeated for every between-hourglass skip connection to fuse the multiscale LR and HR features. Ultimately, the proposed LGSA can enhance HR features to significantly improve dense pest counting.

### 4.5. HR heatmap prediction

To train the HR prediction maps, the HR loss function is defined as follows:

$$\mathcal{L}_T^{HR} = \mathcal{L}_h^{HR} + \lambda_b \mathcal{L}_b^{HR} + \lambda_o \mathcal{L}_o^{HR} \tag{17}$$

$$\mathcal{L}_T = \mathcal{L}_T^{HR} + \mathcal{L}_T^{LR} \tag{18}$$

Here, $\mathcal{L}_T^{HR}$ is equivalent to $\mathcal{L}_T^{LR}$ in Eq. (6) except the HR prediction maps in place of the LR prediction maps. Thus, Eq. (18) indicates that the total loss is the sum of the LR and HR losses. The HR prediction maps are ultimately used for the bounding box prediction, whereas the LR prediction maps are used for the initial local grouping in this study.

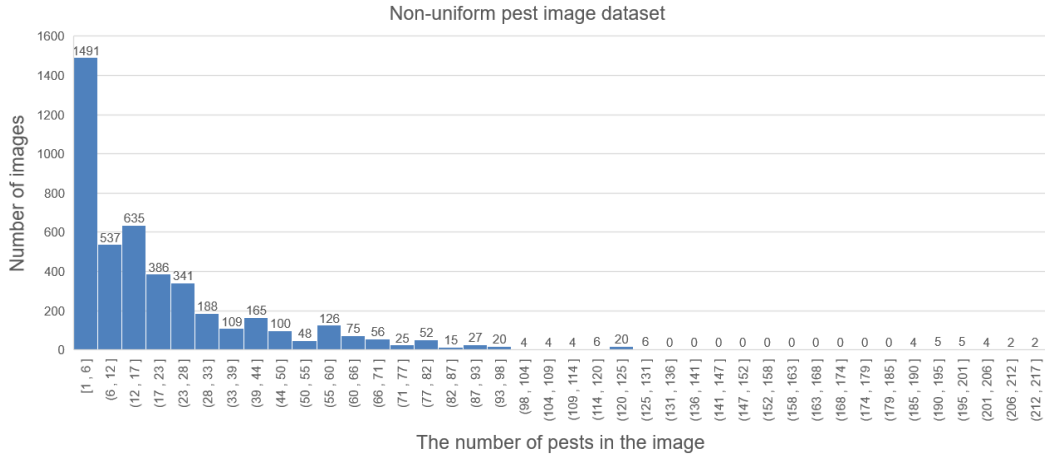

Fig. 6. Histogram of non-uniform pest image dataset.

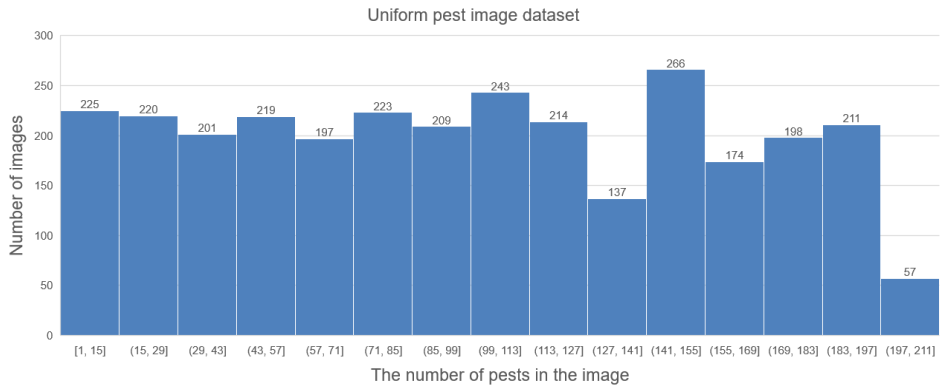

Fig. 7. Histogram of uniform pest image dataset.

## 5. Experimental Results

We used two types of pest image datasets to demonstrate that dense pest counting significantly differs from general object detection, which focuses on sparsely distributed objects. Figs. 6 and 7 show the histograms used to determine the data distributions for the two types of pest datasets, where the horizontal axis represents the number of pests per image and the vertical axis represents the number of images. The brackets on the horizontal axis represent the histogram bins. From the figures, it is clear that the dataset in Fig. 6 is uneven, whereas that in Fig. 7 is almost uniform. In the first dataset, images with pest counts between 1 and 35 account for 80% of the data, while those with pest counts greater than 100 are rare. Consequently, the dataset induces pest-counting models to exhibit a bias toward peak distribution, prompting the question of whether the model can generalize its counting capability across varying pest densities. To test this, we introduced a second dataset with a uniform pest distribution.

The non-uniform and uniform datasets encompass 4,462 and 3,000 images, respectively. The non-uniform dataset contains two types of pests – Spodoptera exigua and Spodoptera litura – whereas the uniform dataset represents only one species – Riptortus clavatus. To construct the datasets, pest specimens were placed in digital traps and photographed using a built-in camera. To reflect the real environment as much as possible, the traps were shaken to occlude or cluster the pest specimens. For training, image flipping was applied for data augmentation, and the images were allocated between training and testing datasets at a 7:3 ratio. Adam [30] was used as the optimizer, and the batch size was set to 4. The number of epochs was 100, and the learning rate was set to 0.001. PyTorch was used as the deep learning framework. The input pest images had a fixed size of $512 \times 512$, and the stride $R$ was set to 4 for downsampling according to the guidelines in [4].

Table 1. Quantitative evaluation of detection-based counting models.

| Datasets | SOTA models | MAE (↓) | RMSE(↓) | AP (↑) |
|---|---|---|---|---|
| Non-uniform dataset | RetinaNet [34] | **4.773** | 9.485 | 0.871 |
| | Faster RCNN [2] | **1.743** | 5.972 | 0.884 |
| | RepPoints [35] | **1.561** | 4.240 | 0.887 |
| | YOLOX [37] | **1.700** | 4.736 | 0.936 |
| | YOLOv7[38] | **1.105** | 2.718 | 0.958 |
| | CenterNet[36] | **0.723** | 2.005 | 0.938 |
| | Multiscale CenterNet[4] | **0.674** | 1.861 | 0.941 |
| | **Proposed LGSA** | 0.647 | 1.589 | **0.967** |
| Uniform dataset | RetinaNet [34] | **39.589** | 52.454 | 0.595 |
| | Faster RCNN [2] | **29.397** | 42.553 | 0.521 |
| | RepPoints [35] | **28.404** | 43.095 | 0.678 |
| | YOLOX [37] | **11.612** | 14.717 | 0.676 |
| | YOLOv7[38] | **4.843** | 16.573 | 0.609 |

|  | CenterNet[36] | **6.015** | 11.031 | 0.653 |
|  | Multiscale CenterNet[4] | **3.334** | 9.719 | 0.693 |
|  | **Proposed LGSA** | **2.538** | **6.666** | **0.920** |

**5.1. Quantitative evaluation of detection-based dense pest counting**

We now describe our evaluation of general object detection models for dense pest counting as well as present comparative results between our LGSA model and existing state-of-the-art (SOTA) models. For quantitative evaluation, three metrics were used: mean absolute error (MAE) [24], root mean squared error (RMSE) [24], and average precision (AP) [2]. Both MAE and RMSE relate to pest-counting accuracy, whereas AP reflects box-detection accuracy. As comparative baselines, we selected SOTA models including RetinaNet [34], RedPoints [35], CenterNet [36], Multiscale CenterNet [4], YOLOX [37], YOLOv7 [38], and Faster R-CNN [2]. Table 1 presents evaluation results achieved by the detection-based SOTA models.

To assess whether general object detection models were effective for dense pest counting, two types of datasets were tested, as mentioned previously. For the non-uniform dataset, general object detection models such as Faster R-CNN, RedPoint, and the YOLO architectures exhibited acceptable counting capabilities. Although these models were initially developed for sparsely distributed objects, they can also be adapted for dense pest counting. Based on these results, it can be concluded that these models are suitable for counting dense pests. However, the performance of these models deteriorated significantly for the uniform dataset. **This result reveals that the performance of general object detection models is significantly influenced by the number of pests, degree of occlusion, and pose variability in images. As with the uniform dataset, when the range of the number of pests is wide, general object detection models experience challenges in localizing densely distributed pests due to significant obstacles such as occlusion and post-variation. That is, the models face difficulties in distinguishing and localizing pests in areas where they are densely distributed or clustered. These results support our motivation that conventional detection-based counting models are not suitable for dense pest counting**.

Table 1 also confirms that the proposed LGSA model outperformed the SOTA models for both datasets, with especially significant improvements on the uniform dataset. **Specifically, the proposed model improved MAE accuracy by approximately 1058.2%, 90.8%, and 31.3% compared with Faster RCNN, YOLOv7, and multiscale CenterNet, respectively. In terms of AP, it also achieved improvements of approximately 35.3%, 33.8%, and 24.6% compared with Faster RCNN, YOLOv7, and multiscale CenterNet, respectively. These improvements represent a notable contribution to the field of dense pest counting**. In addition, Table 1 shows that the CenterNet families – including CenterNet, Multiscale CenterNet, and the proposed LGSA model – tended to exhibit higher overall accuracy than the other baselines. This can be attributed to the relatively high-resolution tensor outputs, which help reduce spatial sampling errors and enhance the detection of dense pests. Compared with the multiscale CenterNet, which applies global attention, the proposed LGSA model exhibited improved performance in terms of all three metrics, indicating that local grouping and scale-guided attention are effective in localizing densely distributed pests. The heat map predicted by the first hourglass served as a good starting point for local feature grouping, which eliminated the need for complex clustering and helped complete the proposed LGSA.

## 5.2. Visualization of pest counting results

Because the proposed LGSA model is an advanced version of the multiscale CenterNet, a comparison of the two models can verify whether LGSA is effective for dense pest counting. Figs. 8 and 9 show detected bounding boxes using the multiscale CenterNet and proposed LGSA models. As shown in Fig. 8, there were numerous pest objects affected by occlusions and pose variations. In contrast, Fig. 9 depicts a relatively small number of pests; however, several pests are clustered with occlusion, which is slightly different from general object detection datasets [39]. The measured and estimated numbers of pests are given above the original and visualized images, respectively. By comparing these predicted numbers, we can observe that the proposed local-attention-based model outperformed the multiscale CenterNet, which only considers global attention. The enlarged parts of the yellow and green boxes are also provided, verifying that LGSA can overcome obstacles such as occlusion, pose variation, and similar appearance in color and texture, which hinder dense pest counting. As shown in the enlarged sections, the multiscale CenterNet failed to detect bounding boxes for occluded pests or pests with pose variations and slightly different colors. Thus, although the multiscale CenterNet exhibited the best performance among the selected SOTA baselines, the LGSA model was more successful in localizing densely distributed and clustered pests accompanied by occlusion and pose variation. These results confirm that our local grouping and scaled-guided attention are highly effective in localizing dense pests and overcoming challenges such as occlusion and pose variability.

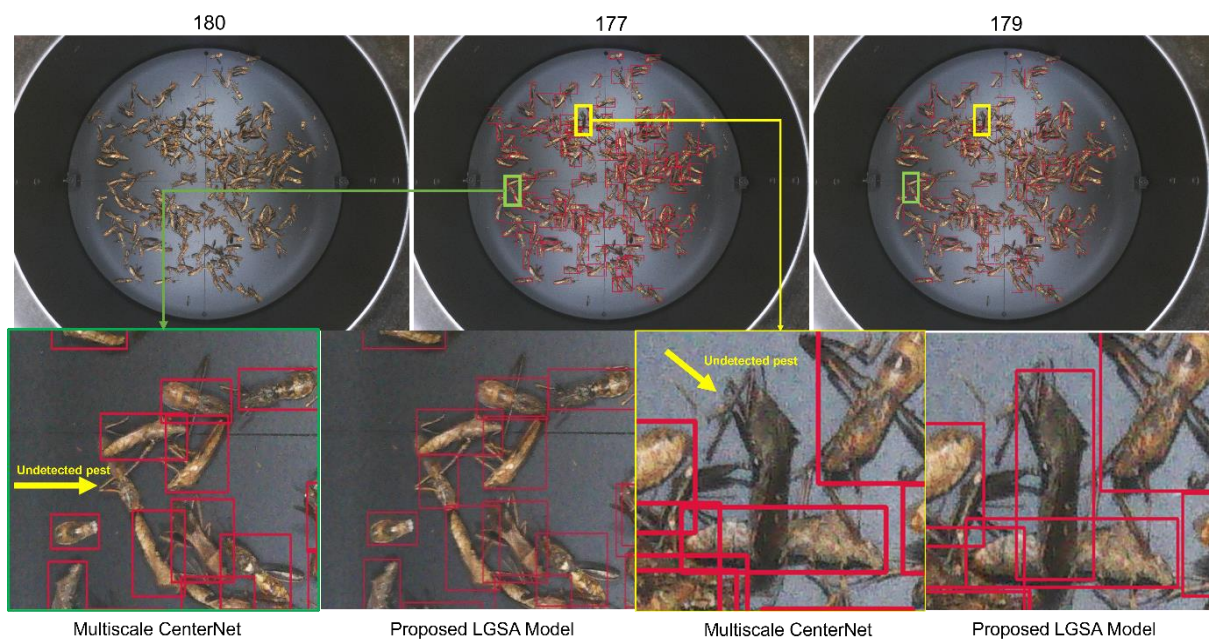

Fig. 8. Visualization of pest detection results for uniform dataset; original image, pests detected with multiscale CenterNet, pests detected with proposed LGSA model (first row from left to right), and enlarged parts of yellow and green boxes (second row).

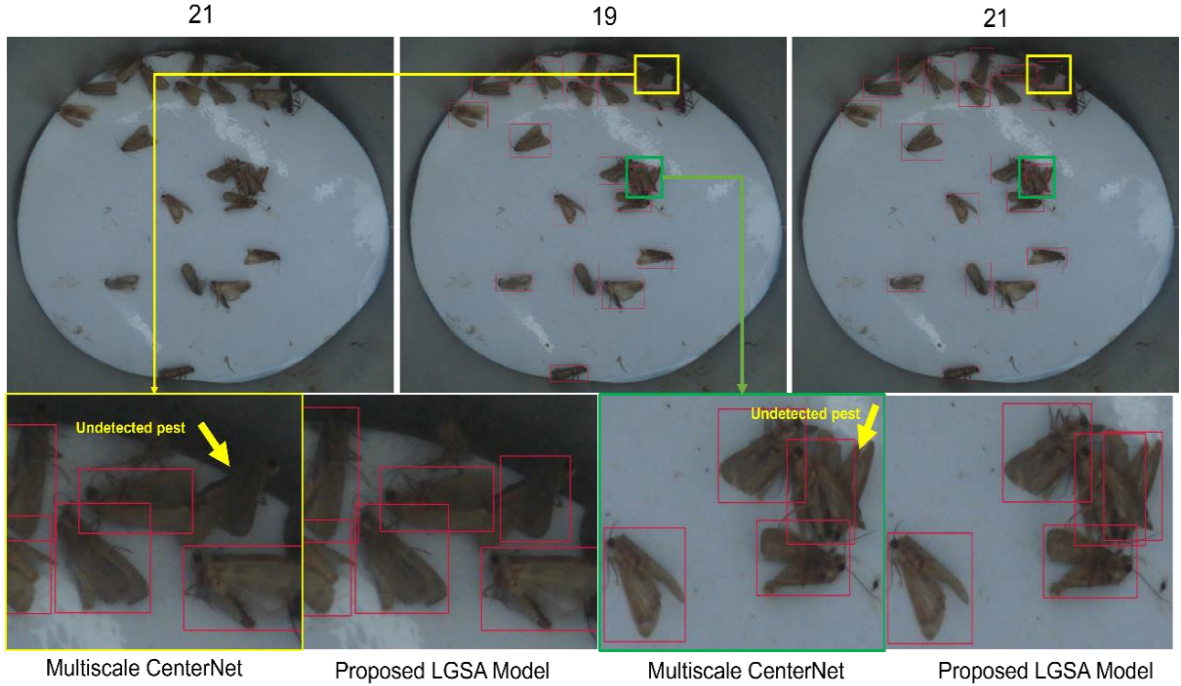

Fig. 9. Visualization of pest detection results for non-uniform dataset; original image, pests detected with multiscale CenterNet, pests detected with proposed LGSA model (first row from left to right), and enlarged parts of yellow and green boxes (second row).

**5.3. Visualization of learnable heatmap and similarity matrix**

In the proposed method, the heatmap information serves as a starting point for local grouping before transforming into a learnable heatmap through pixel attention for discriminative feature learning, as shown in Fig. 4. To verify that the learnable heatmap becomes more discriminative, we visualized heatmaps alongside corresponding bounding box and offset maps, as shown in Figs. 10(b) and (c), where the bounding box map, heatmap, and offset map are listed from left to right. From Fig. 10(c), we can observe that the image contrast of the prediction maps increases further after applying the learnable heatmap transformation, particularly for keypoint areas. That is, the keypoints in all prediction maps, including the heat map, become more noticeable as object features are emphasized more while background features are suppressed. These results confirm the effectiveness of the proposed local method. Therefore, this learnable heatmap transformation helps to localize dense and clustered pests with severe occlusion and wide pose variation, making the proposed LGSA more suitable for dense pest counting. For reference, there are two bounding boxes and offset maps for each image; however, only one of each map is provided, as it is visually more prominent. In addition, the predicted heatmap is presented in grayscale, which is different from its ground-truth map with black and white colors. However, the LR heatmap is only used for local grouping rather than final pest detection. Fig. 10(d) visualizes the similarity matrix. As expected, the matrix is sparse, indicating sufficient separation between object and background locations. Note that image contrast was applied to obtain Fig. 10(d).

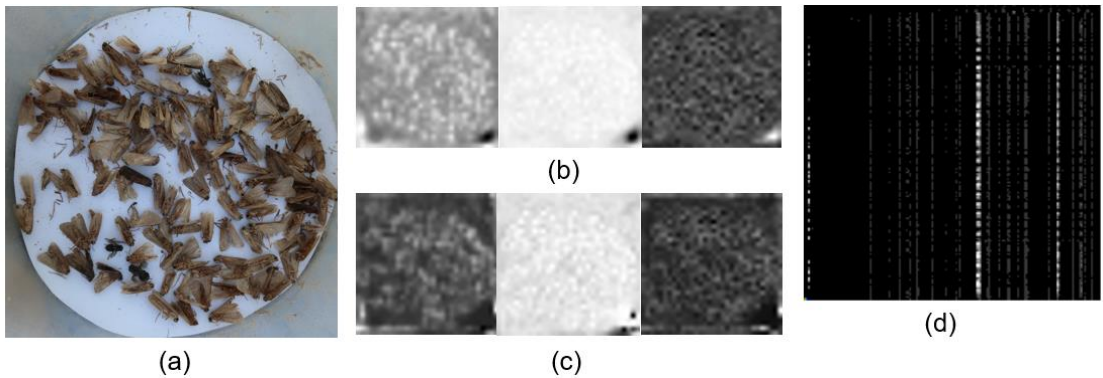

Fig. 10. Visualization results: (a) input image; (b) LR prediction maps before applying the learnable heatmap transformation where the bounding box map, heatmap, and offset map are listed from left to right; (c) LR prediction maps after applying the learnable heatmap transformation where the bounding box map, heatmap, and offset map are listed from left to right; (d) similarity matrix.

**5.4. Discussion on detection- and density-based counting models**

We now verify whether density-based counting models apply to dense pest counting and then discuss detection- and density-based models. Density-based counting models were originally designed for crowd counting. Images of crowds resemble those of dense pests, as both contain numerous objects and occlusions. Therefore, density-based counting models may be more suitable for dense pest counting. Table 2 presents evaluation results of popular density-based pest counting models such as MCNN[25], SANet[26], ICCNet[27], and DM-Count[29]. As observed from the table, density-based counting models are actually slightly less accurate than detection-based counting models for nonuniform datasets. Although DM-Count is comparable to detection-based models, its accuracy is lower than that of the CenterNet family of models. These results indicate that detection-based counting models effectively find bounding boxes from images with fewer than 35 pests, accounting for 80% of the non-uniform dataset. These results empirically offer a rough idea of how many pests a given detection-based counting model can detect. For the uniform dataset, the density-based counting models exhibited better overall performance than the detection-based counting models. In particular, DM-Count and ICCNet achieved considerably better MAE measures than the detection-based methods. The uniform dataset consists of images with numerous severe obstacles, such as occlusion and pose variation, and detection-based models often fail to identify bounding boxes for occluded and clustered objects. In contrast, density-based counting models estimate the density map, which resembles the heatmap, thereby mitigating the aforementioned obstacles and adapting to the dense pet counting task. We therefore conclude that the density-based counting models apply to dense pest counting.

Table 2. Quantitative evaluation of density-based counting models.

| Datasets | SOTA models | MAE (↓) | RMSE(↓) |
|---|---|---|---|
| Non-uniform dataset | MCNN [25] | 4.104 | 7.493 |
| | ICCNet [27] | 9.710 | 11.000 |
| | SANet [26] | 2.114 | 3.740 |
| | DM-Count [29] | 0.888 | 2.013 |
| Uniform dataset | MCNN [25] | 3.659 | 5.054 |
| | ICCNet [27] | 2.212 | 4.064 |
| | SANet [26] | 5.174 | 7.258 |
| | DM-Count [29] | 2.393 | 3.712 |

This study was conducted to develop a new detection-based counting model for dense and clustered pests as an alternative to density-based counting models. Unlike density-based models such as DM-Count and ICCNet, detection-based models not only count the number of pests but also localize pests. In other words, the two types of counting models have slightly different objectives. Overall, the proposed LGSA model outperforms density-based models on the nonuniform dataset and it is comparable to the performance of DM-Count and ICCNet on the uniform dataset. In particular, the proposed model significantly outperformed detection-based models on both datasets, yielding a notable contribution to dense pest counting. However, to complement the proposed model in future studies, we will consider a hybrid structure that combines the two types of counting models to adapt to the number of pests. In particular, a density map will be combined with a heat map to enhance predictive accuracy.

## 6. Conclusion

This study examines the task of counting and identifying densely distributed pests trapped by digital traps, which presents challenges such as severe occlusion, wide pose variation, and similar appearance in colors and textures. To address these problems, we designed the LGSA model as a novel dense counting model, representing a simple yet highly effective approach for separating object features from background features and making them more discriminative based on learnable heat map transformations. Our experimental results show that the proposed model outperforms existing SOTA models by a significant margin, particularly on the uniform dataset, representing a notable improvement in dense pest counting. In particular, the proposed model has proven to be highly effective in overcoming challenges such as occlusion and pose variability stemming from local feature grouping and discriminative feature attention learning. In future studies, we intend to advance our dense counting model by considering a hybrid approach that combines density and heat maps to adapt to the number of pests and further enhance predictive accuracy.

**Funding**

This work was supported by the Cooperative Research Program for Agriculture Science and Technology Development (Grant No. PJ016303), the National Institute of Crop Science (NICS), Rural Development